\documentclass[11pt]{article}

\def\0{{\bf 0}}
\def\1{{\bf 1}}

\newtheorem{innercustomgeneric}{\customgenericname}
\providecommand{\customgenericname}{}

\newcommand{\tabref}[1]{Table~\ref{#1}}

\newcommand{\figref}[1]{Fig.~\ref{#1}}

\renewcommand{\cite}{\citep}

\usepackage[final]{acl}

\usepackage{times}
\usepackage{latexsym}
\usepackage{booktabs}
\usepackage{algorithm2e}
\usepackage{subfigure}
\usepackage{multirow}
\usepackage{amsfonts}
\usepackage{amsmath}
\usepackage{graphicx}
\usepackage{subcaption}
\usepackage{mdframed}
\usepackage{float}
\usepackage{algpseudocode}

\usepackage{listings}
\usepackage{xcolor}
\usepackage{paralist}

\usepackage[T1]{fontenc}
\usepackage[utf8]{inputenc}

\usepackage{microtype}

\usepackage{inconsolata}

\usepackage{graphicx}

\title{Compressing Sequences in the Latent Embedding Space:\\$K$-Token Merging for Large Language Models}

\author{
Zihao Xu\textsuperscript{\ensuremath{\diamondsuit}}\thanks{Work done at AWS AI Labs.}, 
John Harvill\textsuperscript{\ensuremath{\heartsuit}}\footnotemark[1],
Ziwei Fan\textsuperscript{\ensuremath{\spadesuit}}, 
Yizhou Sun\textsuperscript{\ensuremath{\clubsuit\dagger}}, \\
\textbf{Hao Ding}\textsuperscript{\ensuremath{\spadesuit}},
\textbf{Hao Wang}\textsuperscript{\ensuremath{\diamondsuit\triangle}}, 
\\
\textsuperscript{\ensuremath{\diamondsuit}}Rutgers University,
\textsuperscript{\ensuremath{\spadesuit}}AWS AI Labs,
\textsuperscript{\ensuremath{\clubsuit}}Amazon, 
\textsuperscript{\ensuremath{\heartsuit}}Mistral AI\\
\textsuperscript{\ensuremath{\dagger}}University of California Los Angeles, \textsuperscript{\ensuremath{\triangle}}University of Illinois Urbana-Champaign\\
\texttt{zihao.xu@rutgers.edu}
}

\usepackage{url}

\begin{document}
\maketitle
\begin{abstract}
Large Language Models (LLMs) incur significant computational and memory costs when processing long prompts, as full self-attention scales quadratically with input length. Token compression aims to address this challenge by reducing the number of tokens representing inputs. However, existing prompt-compression approaches primarily operate in token space and overlook inefficiencies in the \textbf{latent embedding space}. In this paper, we propose $K$-Token Merging, a latent-space compression framework that merges each contiguous block of $K$ token embeddings into a single embedding via a lightweight encoder. The compressed sequence is processed by a LoRA-adapted LLM, while generation remains in the original vocabulary. Experiments on structural reasoning (Textualized Tree), sentiment classification (Amazon Reviews), and code editing (CommitPackFT) show that $K$-Token Merging lies on the Pareto frontier of performance vs. compression, achieving up to \textbf{75\%} input length reduction with minimal performance degradation. On a Qwen Coder 7B model, 2-Token Merging reduces latency by 20.9\% and improves throughput by 27.8\% with only a 1.1\% relative increase in perplexity. Code is available at \url{https://github.com/shsjxzh/K-Token-Merging}.

% Large Language Models (LLMs) incur significant computational and memory costs when processing long prompts. 

% While prior work on prompt compression has primarily relied on lossy methods that discard tokens, these approaches fail on information-dense tasks where all input tokens are critical. In this paper, we propose a token compression framework that operates in the \textbf{latent embedding space}, compressing sequences of multiple tokens into a single, compact representation. Our method employs a lightweight encoder to merge $k$-token embeddings into one, and a LoRA-adapted LLM to process the compressed sequence. We demonstrate strong results across synthetic (Tree Classification), natural language (Amazon Reviews), and code (CommitPackFT) benchmarks, achieving up to \textbf{75\% token reduction} with minimal or no performance degradation.
\end{abstract}

\section{Introduction}

\begin{figure}[!htbp] %{wrapfigure}{r}{0.45\columnwidth}
% \vskip -0.14in
\centering
\includegraphics[width=0.95\columnwidth]{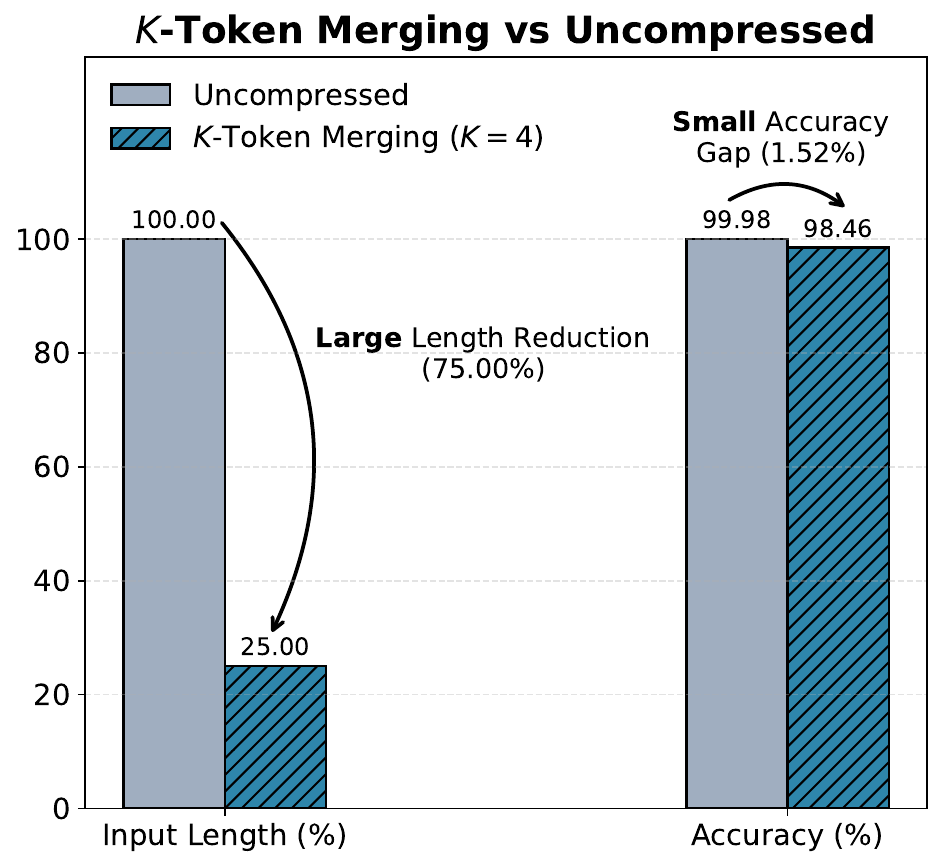}
% \vskip -0.10in
\caption{Our $K$-Token Merging method ($K = 4$) achieves a $75\%$ reduction in input length with only a $1.52\%$ drop in accuracy on the Textualized Tree benchmark, demonstrating that it exploits redundancy in the latent embedding space while preserving high performance. See the ``Experiments'' section for details.}
\label{fig:teaser}
\vskip -0.15in
\end{figure}

\begin{figure*}[!t]
\vskip -0.2cm
\begin{center}
\includegraphics[width=0.95\linewidth]{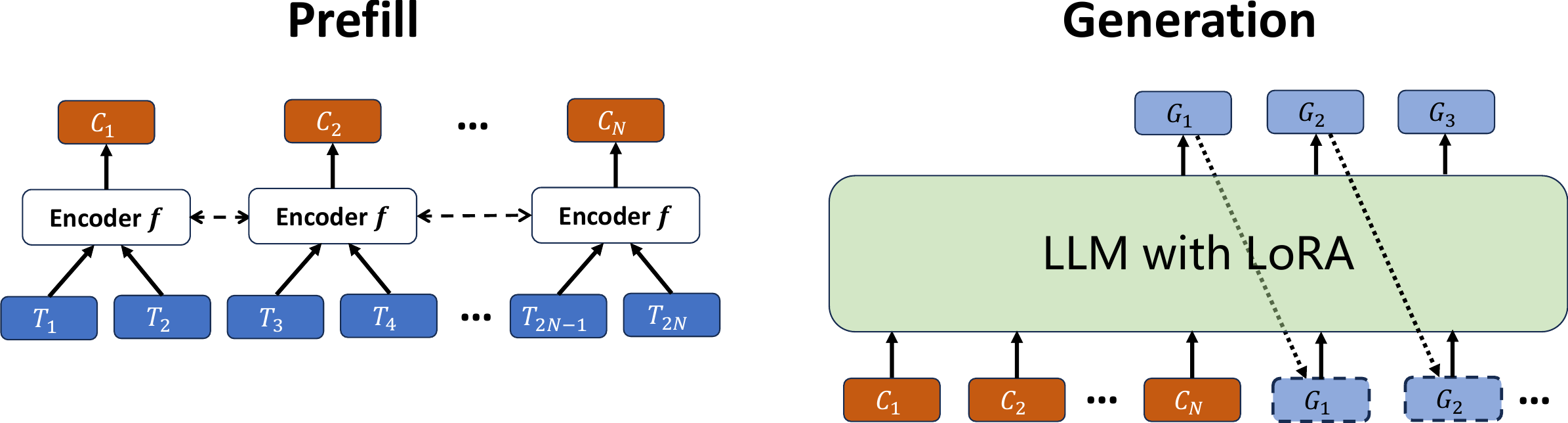}
\caption{\textbf{Model Structure for $K$-Token Merging Model (Case $K=2$).}
\textbf{Left:} During the prefill stage, the encoder $f$ takes each $K$ consecutive input tokens and produces a single compressed token embedding. Here, $T_i$ denotes the original input tokens and $C_i$ denotes the resulting compressed tokens.
\textbf{Right:} During the generation stage, the LLM outputs \emph{original} (uncompressed) tokens. Each newly generated token is appended to the mixed compressed/uncompressed prefix, after which standard auto-regressive generation continues. Here, $G_i$ denotes the generated uncompressed tokens.}
\label{fig:model}
\end{center}
\vskip -0.3cm
\end{figure*}

We have witnessed tremendous progress in Large Language Models (LLMs)~\citep{llm_survey}. Since the advent of ChatGPT~\citep{gpt3}, LLMs have become deeply integrated into many aspects of daily life -- from AI-enhanced customer support~\citep{shareef2024enhancing, scatolin2026stellar} and game NPCs~\citep{cox2023conversational, christiansen2024exploring} to agentic systems capable of data analysis~\citep{tang2025llm} and complex software generation~\citep{joel2024survey, jiang2026survey}. These increasingly sophisticated applications demand ever-longer input contexts, which sharply escalate both computational and memory costs: with softmax attention~\citep{vaswani2017attention}, the memory and compute requirements of LLMs grow \textbf{quadratically} with input length.

A natural direction to mitigate this challenge is \emph{token compression} ~\citep{li2025prompt}. Token compression aims to reduce the input length by representing prompts with fewer tokens. Existing approaches largely fall into two categories.

The first category, \textbf{hard prompt compression}~\citep{li2023compressing, llmlingua, longllmlingua, liskavets2025prompt}, decreases token count by dropping or summarizing tokens. While effective for tasks where key information is sparse, these methods often fail on information-dense tasks such as document revision, mathematical reasoning, symbolic computation, or code translation--settings in which every token may carry essential information.

% Model Structure for $K$-Token Merging Model when $K=2$. \textbf{Left:} In the prefilling stage, our encoder $f$ takes $K$ consecutive tokens as input and output one compressed token embedding. Here, $T_i$ denotes input tokens, and $C_i$ denotes compressed tokens. \textbf{Right:} In the generation stage, our LLM outputs \emph{original} (uncompressed) tokens. These newly produced tokens are appended to the mixed compressed/uncompressed prefix and then we do the traditional auto-regression of LLM to keep doing generation. Here, $G_i$ denotes generated uncompressed tokens.

The second category, \textbf{soft prompt compression}~\citep{gist, LTSC, gao2024uniicl}, draws inspiration from soft prompts~\citep{lester2021power} and learns new tokens or adapts model parameters to produce more compact representations. These methods have gained increasing attention because, empirically, they preserve more information than hard methods while still offering substantial compression~\citep{LTSC, gao2024uniicl}. However, existing methods focus almost exclusively on reducing the number of \emph{tokens}, overlooking a major source of redundancy: \emph{the embedding space itself}.

Consider the \textsc{Qwen~2.5} model~\citep{qwen2.5}, which has a vocabulary size of $151{,}936$. Each token consumes approximately $896 \times 32 = 28{,}672$ bits, whereas the theoretical minimum needed to identify one token from its vocabulary is only $\log_2(151{,}936) \approx 18$ bits. For a $K$-gram, the minimum grows to merely $\log_2(151{,}936^K) \approx 18K$ bits. This enormous gap highlights significant inefficiency in current embeddings and indicates substantial room for improving input representation.

In this paper, we propose \textbf{$K$-Token Merging}, a new soft prompt compression method that directly targets redundancy in the \emph{latent embedding space}. Our key idea is to use a learned encoder to compress a sequence of $K$ consecutive tokens ($K$-grams) into a single embedding. In other words, we represent multiple tokens with one embedding \textbf{on the input side only}. We then finetune the LLM with LoRA to adapt to these compressed embeddings, enabling strong downstream performance.

As illustrated in Figure~\ref{fig:teaser}, our approach achieves high compression ratios with minimal performance degradation (see the ``Experiments'' section for details).

Importantly, our method is fundamentally different from expanding the vocabulary with $K$-grams. Vocabulary expansion requires representing multi-token sequences as single tokens \emph{on both the input and output sides}. This leads to an explosion of low-frequency ``tail'' tokens, which severely complicates optimization and becomes quickly infeasible for large $K$. For example, the number of unique 4-grams can exceed the size of the original vocabulary by $29\times$~\citep{yu2025scaling}, resulting in prohibitive storage overhead. Furthermore, since each $K$-gram is assigned a fixed embedding, such approaches cannot generalize to unseen $K$-grams during inference, sharply limiting their applicability. In contrast, our encoder-based approach avoids tail-token explosion, supports large $K$ values without increasing vocabulary size, and generalizes naturally to unseen token combinations.

Our main contributions can be summarized as follows:

\begin{figure*}[!htbp]
\begin{center}
\includegraphics[width=0.95\linewidth]{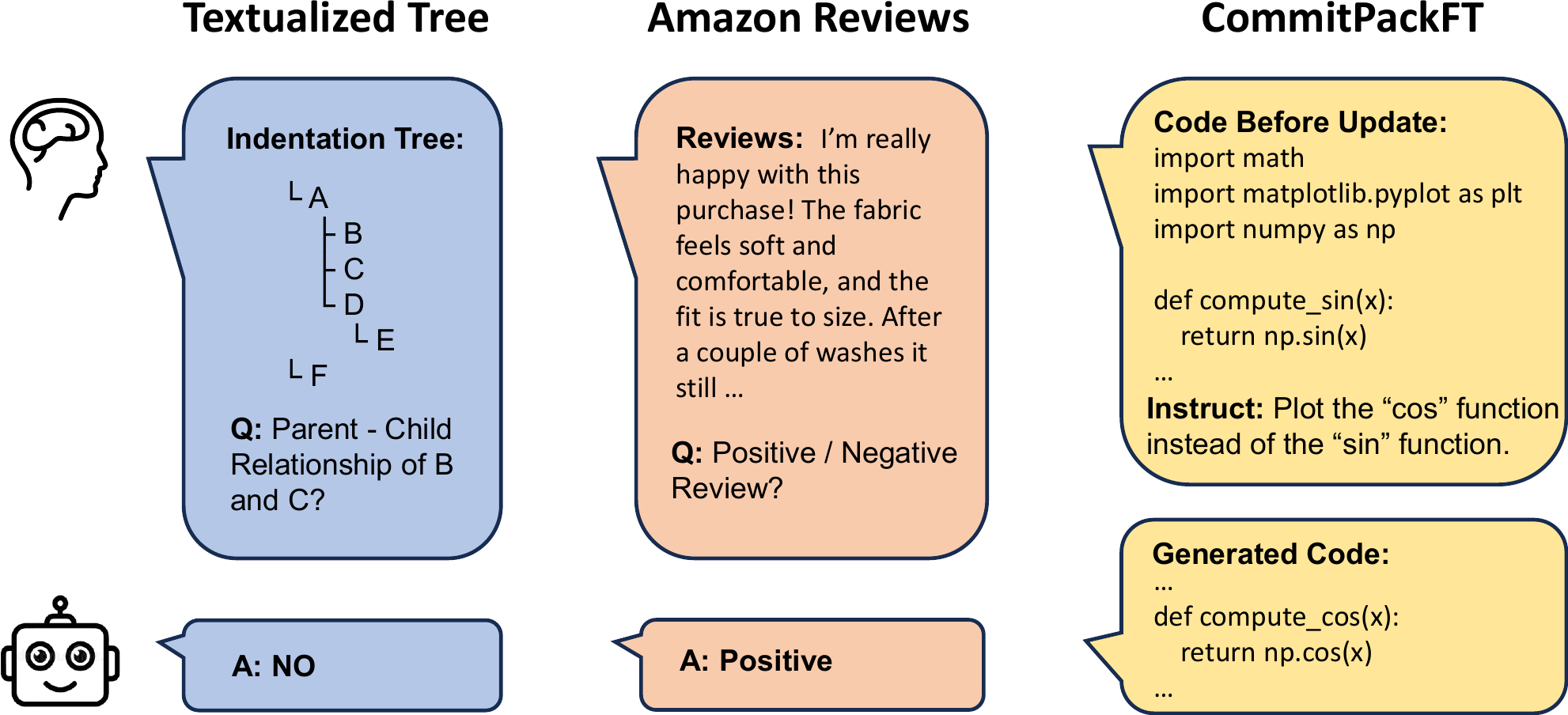}
\caption{\textbf{Datasets \& Tasks.}
\textbf{Left: Textualized Tree.} Given a textualized indentation tree, the LLM determines whether two nodes have a parent - child relationship.
\textbf{Middle: Amazon Reviews.} The LLM performs sentiment classification to judge whether a product review is positive or negative.
\textbf{Right: CommitPackFT.} Given code and an update instruction, the LLM needs to output the modified code that follows the instruction.}
\label{fig:dataset_visualization}
\end{center}
\vskip -0.3cm
\end{figure*}

\begin{compactitem}
    \item We introduce \textbf{$K$-Token Merging}, a novel framework for latent-space token sequence compression.
    \item We present a training recipe for adapting LLMs to such compressed inputs.
    \item We show experimentally that our approach achieves up to \textbf{75\% token compression} with minimal performance degradation across diverse tasks. In scenarios where output length is much shorter than input length, this corresponds to an estimated \textbf{94\%\footnote{With K=4, input length is reduced by 75\%. Prefill attention cost scales as O(n²), so compressed prefill requires only $0.25^2 = 6.25\%$ of the original FLOPs. When output length $M \ll$ input length $KN$, total FLOPs $\approx 6.25\%$ of uncompressed, yielding $\sim94\%$ reduction.} reduction in computation}. Additional component ablations and experiments with a 7B model validate the design and demonstrate that its runtime benefits persist at a larger scale.
\end{compactitem}

\section{Related Work}

Token compression~\citep{li2025prompt} has been extensively studied since the introduction of the Transformer architecture~\citep{vaswani2017attention}. Existing approaches can be grouped into \emph{hard prompt compression} and \emph{soft prompt compression}.

\paragraph{Hard prompt compression.}
Hard compression methods~\citep{li2023compressing, llmlingua, chuang2024learning, longllmlingua, jung2024discrete, taco-rl, liu2023tcra, liskavets2025prompt} reduce input length by dropping or summarizing tokens. For example, SelectiveContext~\citep{li2023compressing} removes tokens deemed uninformative based on self-information, and LLMLingua2~\citep{llmlingua} trains a classifier to identify redundant tokens. However, these methods often suffer significant performance degradation on information-dense tasks, as even a small number of omitted tokens may contain essential context. 

In contrast, our method never removes tokens; it re-embeds all tokens into compressed embeddings. This enables substantially higher compression rates while mitigating information loss. Furthermore, our technique is compatible with hard compression and can be applied subsequently, making it complementary rather than competing.

\paragraph{Soft prompt compression.}
Soft compression methods~\citep{bolya2022token,gist, chevalier2023adapting, ge2023context, xrag, LTSC, gao2024uniicl} learn new tokens or adapt LLM parameters to produce compact input representations. For instance, Gist \cite{gist} compresses instructions into a small set of meta-tokens prepended to the input, enforcing restricted attention patterns to ensure that later tokens cannot be attended beyond the meta-tokens. LTSC \cite{LTSC} discovers frequently occurring token subsequences and replaces them with shorter learned meta-tokens, analogous to classical file-compression algorithms.

However, these methods operate entirely at the \emph{token level}. In contrast, our approach performs compression directly in the \emph{latent embedding space}, targeting inefficiencies in token embeddings themselves rather than the token sequence. This distinction allows our framework -- though still a form of soft compression -- to achieve a substantially more favorable balance between performance and compression ratio than prior approaches.

\section{Method}

% \begin{table*}
% \setlength{\tabcolsep}{5pt}
% \caption{Accuracy (\%), Length Reduction Ratio (\%), and P--L $F_1$ on Textualized Tree. We \textbf{bold} the best result and \underline{underline} the second-best result for each metric. Our method achieves both the best and second-best results on Length Reduction Ratio and P--L $F_1$, while achieving the second-best Accuracy, surpassed only by the Uncompressed baseline.}
% \label{tab:tree_classify}
% \vskip 0.15cm
% \centering
% \begin{small}

% \begin{tabular}{ccccccccc}
% \toprule
% \multirow{2}{*}{Method} 
%   & \multirow{2}{*}{Uncompressed} 
%   & \multirow{2}{*}{SelectiveContext} 
%   & \multirow{2}{*}{LLMLingua2\footnotemark[1]} 
%   & \multirow{2}{*}{LTSC\footnotemark[1]}
%   & \multicolumn{3}{c}{\textbf{$K$-Token Merging (Ours)}} \\
% \cmidrule(lr){6-8}
%   &   &   &   &   
%   & 2-Token & 3-Token & 4-Token \\
% \midrule
% \emph{Accuracy} 
%   & \textbf{99.97} & 90.43 & 82.17 & 99.68 
%   & \underline{99.91} & 98.63 & 98.38 \\
% \emph{Length Reduction}  
%   & 0.0 & 52.5 & 27.0 & 27.1 
%   & 50.0 & \underline{66.7} & \textbf{75.0} \\
% \emph{P--L $F_{1}$}  
%   & 0.000 & 0.664 & 0.406 & 0.426 
%   & 0.666 & \underline{0.796} & \textbf{0.851} \\
% \bottomrule
% \end{tabular}

% \end{small}
% \end{table*}

\begin{table*}
\setlength{\tabcolsep}{5pt}
\caption{Accuracy (\%), Length Reduction Ratio (\%), and P--L $F_1$ on Textualized Tree. We \textbf{bold} the best result and \underline{underline} the second-best result for each metric. Our method achieves both the best and second-best results on Length Reduction Ratio and P--L $F_1$, while achieving the second-best Accuracy, surpassed only by the Uncompressed baseline.}
\label{tab:tree_classify}
\vskip 0.15cm
\centering
\begin{small}

\begin{tabular}{ccccccccc}
\toprule
\multirow{2}{*}{Method} 
  & \multirow{2}{*}{Uncompressed} 
  & \multirow{2}{*}{SelectiveContext} 
  & \multirow{2}{*}{LLMLingua2} 
  & \multirow{2}{*}{LTSC}
  & \multicolumn{3}{c}{\textbf{$K$-Token Merging (Ours)}} \\
\cmidrule(lr){6-8}
  &   &   &   &   
  & 2-Token & 3-Token & 4-Token \\
\midrule
\emph{Accuracy} 
  & \textbf{99.98}$^{\pm 0.01}$ & 90.33$^{\pm 0.07}$ & 82.17\footnotemark[1] & 99.68\footnotemark[1] 
  & \underline{99.79}$^{\pm 0.22}$ & 98.73$^{\pm 0.55}$ & 98.46$^{\pm 0.30}$ \\
\emph{Length Reduction}  
  & 0.0 & 52.5 & 27.0 & 27.1 
  & 50.0 & \underline{66.7} & \textbf{75.0} \\
\emph{P--L $F_{1}$}  
  & 0.000 & 0.664 & 0.406 & 0.426 
  & 0.666 & \underline{0.796} & \textbf{0.851} \\
\bottomrule
\end{tabular}

\end{small}
\end{table*}

\subsection{Model Structure}
Our insight is to exploit inefficiency in token-level representations. We introduce a lightweight encoder $f$ that compresses every contiguous block of $K$ tokens into a single compressed token embedding. The overall architecture is illustrated in \figref{fig:model}, using the 2-Token Merging model as an example.

For our $K$-Token Merging model, we assume the total number of input tokens be $KN$. If necessary, we append padding tokens to ensure that the input length is always a multiple of $K$. We describe the model’s workflow in two stages: the \textbf{prefill} stage and the \textbf{generation} stage.

\textbf{Prefill.} In the prefill stage, for each block $\{T_{iK+j}\}_{j=1}^{K}, i \in \{0,\dots,N-1\}$,
the encoder $f$ takes as input their original embeddings from the model's embedding table $Emb$ and outputs a single compressed embedding $C_i$:
\begin{equation}
    C_i = f\!\left( Emb(T_{iK+1}), \dots, Emb(T_{iK+K}) \right).
\end{equation}

This compression is computationally inexpensive because: (i) the encoder is a small multi-layer perceptron (MLP) of roughly 50 MB, and (ii) compressed embeddings for frequently occurring $K$-grams may be cached.

\textbf{Generation.} During generation, the model always outputs \emph{original} (uncompressed) tokens. These newly produced tokens are appended to the mixed compressed/uncompressed prefix and autoregressively processed by the LLM.

\subsection{Objective Function}

We finetune the LLM using a LoRA adapter jointly with the encoder $f$. Importantly, the training objective is evaluated only on positions corresponding to uncompressed (original) tokens -- e.g., the generated tokens $G_1, G_2, G_3, \dots$ in \figref{fig:model}. Let the full sequence after inserting compressed tokens be
\[
    X = (C_1, \dots C_N, G_1, \dots G_M),
\]
where $M$ is the number of generated tokens. Let $\mathcal{U} = \{N + 1,\dots,N + M\}$ denote the index set of uncompressed token positions. The LLM defines next-token probabilities $p_\theta(\cdot \mid X_{<t})$. The training loss is the negative log-likelihood restricted to uncompressed targets:
\begin{equation}
    \mathcal{L}(\theta, f)
    = - \sum_{t \in \mathcal{U}}
        \log p_\theta\!\left( X_t \,\middle|\, X_{<t} \right).
\end{equation}

This objective ensures that the model learns to interpret compressed embeddings while maintaining generation quality on the original vocabulary.

\footnotetext[1]{Results come from \citet{LTSC}.}

\subsection{Compressed Embedding Initialization}

The initialization strategy for compressed embeddings has a significant impact on the convergence speed of training. Empirically, we find that initializing the embedding as the average (mean pooling) of the $K$ original token embeddings leads to substantially faster convergence compared to random initialization.

We design a dedicated encoder architecture that naturally produces such an initialization. The encoder computes a residual combination of (i) the mean of the $K$ original embeddings and (ii) the output of a small MLP. The MLP weights are initialized close to zero so that, at the beginning of training, its contribution is negligible. Consequently, the encoder initially outputs a compressed embedding that closely matches the average-pooled embedding of the $K$ tokens, providing a stable and semantically grounded starting point for optimization.

The pseudo code for this \textit{Average-Initialized Encoder} is shown in Alg.~\ref{alg:avg_encoder}.

\section{Experiments}

In this section, we evaluate our method across three tasks, analyze its components and embedding initialization strategy, and present a qualitative case study.

\begin{figure*}[!htbp]
\begin{center}
% \vskip 0.2cm
\subfigure[Textualized Tree]{
\includegraphics[width=0.316\linewidth]{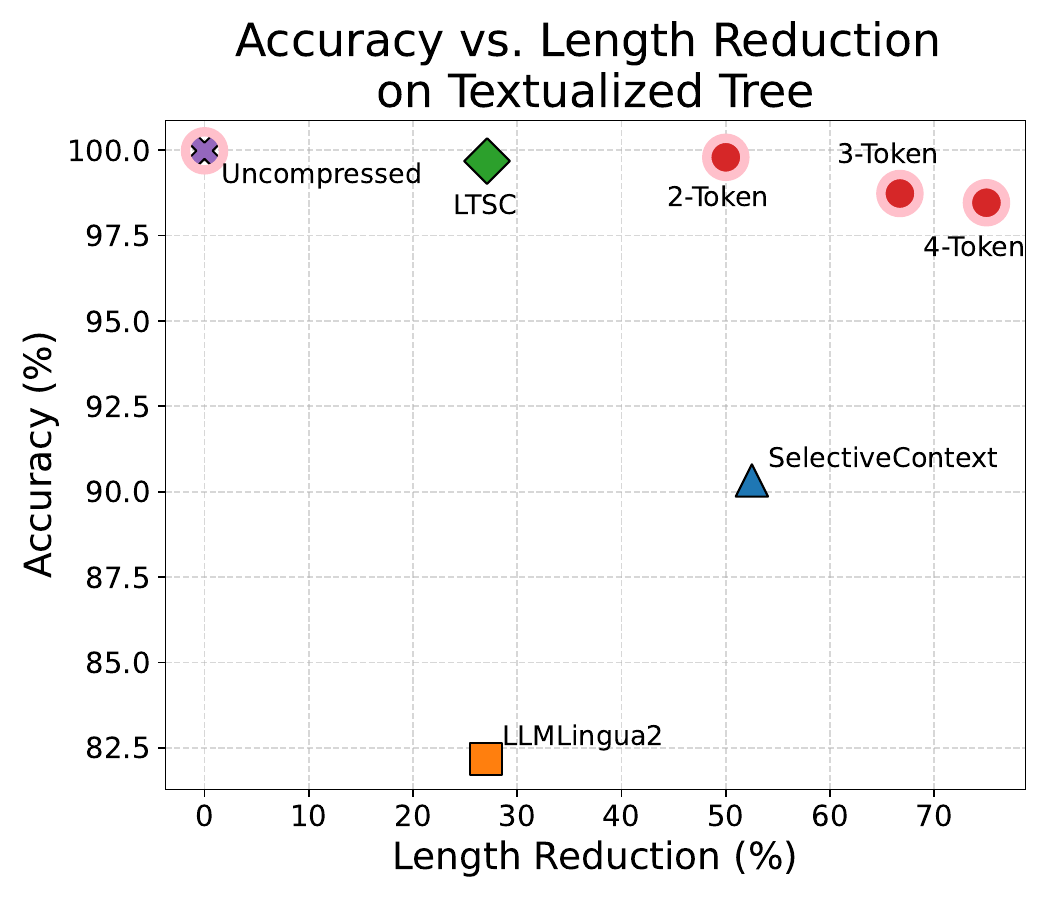}
}
\subfigure[Amazon Reviews]{
\includegraphics[width=0.316\linewidth]{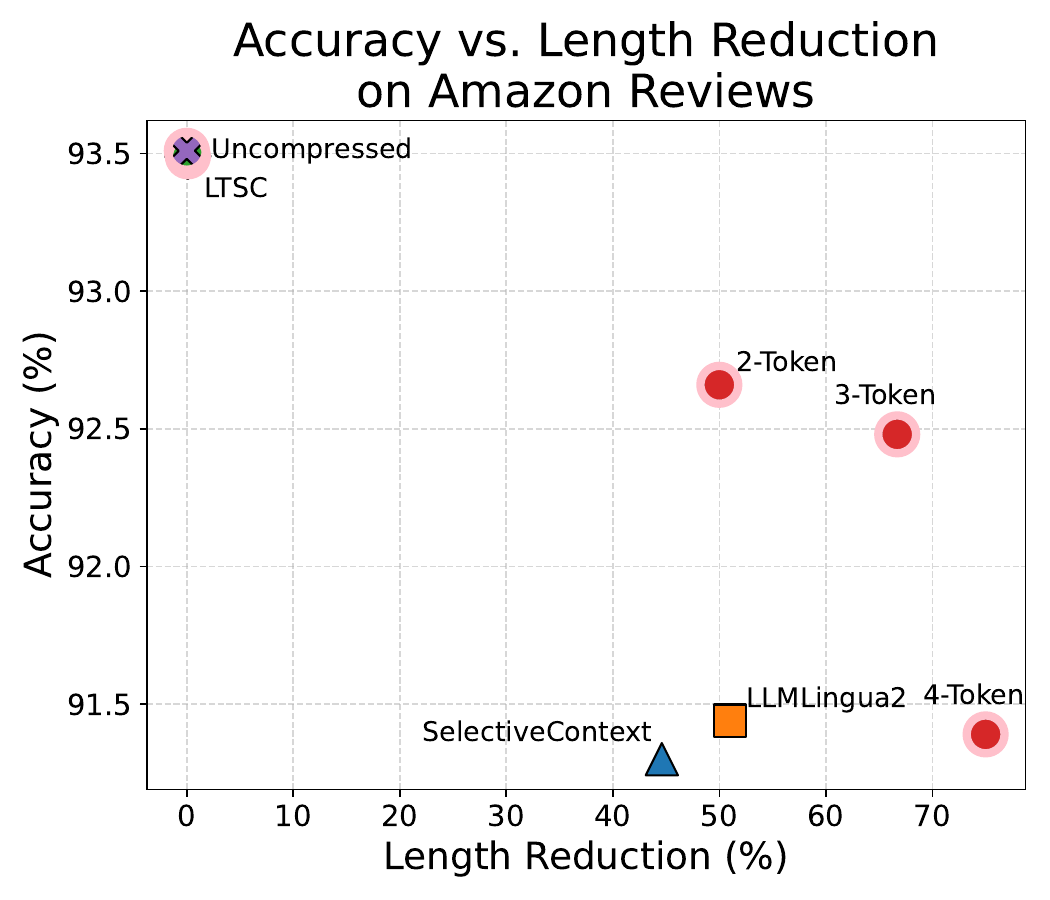}
}
\subfigure[CommitPackFT]{
\includegraphics[width=0.316\linewidth]{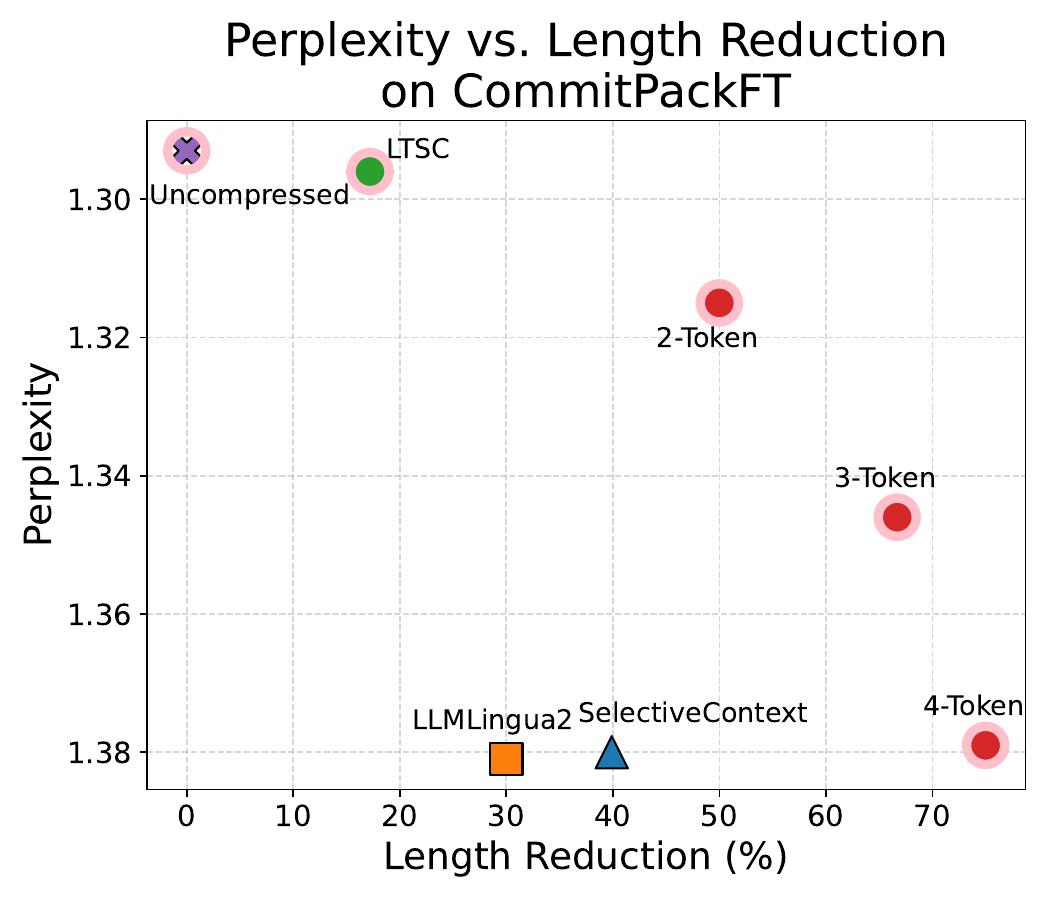}
}
\vskip -0.1cm
\caption{
\textbf{Performance Score (Accuracy / Perplexity) vs. Length Reduction Ratio} on three datasets: (a) Textualized Tree, (b) Amazon Reviews, and (c) CommitPackFT. A higher Performance Score / Length Reduction Ratio indicates better performance; therefore, points located toward the upper-right region of the plot are preferred. Pareto-optimal points are marked with hollow \textcolor{pink}{pink} circle  markers (\textcolor{pink}{$\circ$}). Our method, $K$-Token Merging with $K \in \{2,3,4\}$, is highlighted in \textcolor{red}{red}. As shown, our approach lies on the Pareto-optimal frontier across all three datasets. For consistency, panel (c) (CommitPackFT) uses a y-axis (perplexity) that increases from top to bottom.}
\vskip -0.2cm
\label{fig:expr_result}
\end{center}
\end{figure*}

\subsection{Baselines}
We select two representative hard prompt compression methods, SelectiveContext \citep{li2023compressing} and LLMLingua2 \citep{llmlingua}, along with a recent soft prompt compression approach, LTSC \citep{LTSC}, as our baselines. For comparison, we also report results from the uncompressed model.

\subsection{Datasets \& Tasks}
% need a figure if possible
We use three datasets, each designed for a unique task. 

\textbf{\emph{Textualized Tree}}~\citep{LTSC} is a synthetic dataset consisting of textualized indentation trees, as shown in \figref{fig:dataset_visualization} (left). We design a classification task on this dataset to evaluate whether our method preserves most of the information after prompt compression. Specifically, we randomly select a pair of nodes in the tree and ask the LLM to classify whether a parent-child relationship exists between them. The model outputs ``true'' if such a relationship exists, and ``false'' otherwise. Since any node pair can be selected, the compressed token embeddings must preserve complete structural information to maintain high classification accuracy. This dataset includes 2.45 million training samples and 50 thousand test samples, with each tree containing up to 150 nodes.
% Classification accuracy is used as the evaluation metric.

\textbf{\emph{Amazon Reviews}}~\citep{AmzReview} is a dataset comprising user reviews, item metadata, and user-item interactions from Amazon. We formulate a sentiment classification task to test whether the compressed embeddings perform well on natural language, as shown in \figref{fig:dataset_visualization} (middle). Given a user review (e.g., ``This legging is amazing''), the model is expected to classify the sentiment as either ``positive'' or ``negative''. Reviews with a rating above 3.5 are labeled as positive, and those below or equal to 3.5 as negative. We use the ``Amazon Fashion'' category, consisting of 2.45 million training samples and 50 thousand test samples.
% Classification accuracy serves as the evaluation metric.

\textbf{\emph{CommitPackFT}}~\citep{commitPack} is a Git commit dataset covering 350 programming languages. Each sample contains the code before an update, the corresponding commit message, and the updated code. We define a code update task where the model receives the code before the update and the commit message as input, and is expected to generate the updated code, which is shown in \figref{fig:dataset_visualization} (right). We use the CommitPackFT subset, which consists of high-quality filtered data from the original CommitPack dataset. We further focus our experiments on the Python category, which includes 50.4 thousand training samples and 5 thousand test samples. 
% Perplexity on the test set is used as the evaluation metric.

\subsection{Evaluation Metrics}

% For each method, we report the model performance with the compressed prompts and the average length reduction ratio. For the datasets of Texturalized Tree and Amazon Reviews, we use classification accuracy as our performance metric. For CommitPackFT, we use the perplexity as the performance metric.

% However, it is common that for two methods, one methods may have better length reduction ratio while others may have better performance. To quantitatively evaluate methods, we propose Peformance-Length Reduction $F_1$ score (P-L $F_1$), a harmonic mean of the performance and length reduction rate, which is adopted from the classical $F_1$ score.
% The formula for calculating the P-L $F_1$ can be described as follows:

% \begin{equation}
%     \text{P-L}\ F_1 = \frac{2 \cdot {P} \cdot {L}}{{P} + {L}}
% \end{equation}

% Here, $P$ denotes the performance score normalized to $[0, 1]$, and $L$ denotes the length reduction ratio. For Texturalized Tree and Amazon Reviews, we directly use the classification accuracy as $P$, and for CommitPackFT, we use the relative perplexity ratio. Let $\text{ppl}_i$ denotes the perplexity of method $i$, and $\text{ppl}_{\text{min}} = \min\limits_{i} \text{ppl}_i$. The relative perplexity ratio $P_i$ can be described as:
% \begin{equation}
%     P_i = \frac{\text{ppl}_{min}}{\text{ppl}_i},
% \end{equation}
% which is a normalized performance metric falling in $[0,1]$.

For each method, we report both the model performance with compressed prompts and the average length reduction ratio. For the Textualized Tree and Amazon Reviews datasets, we use classification accuracy as the performance metric, while for CommitPackFT, we use perplexity.

\begin{table*}
\setlength{\tabcolsep}{2pt}
\caption{Accuracy (\%), Length Reduction Ratio (\%), and P--L $F_1$ on Amazon Reviews. We \textbf{bold} the best result and \underline{underline} the second-best result for each metric. Our method achieves the best and second-best results on P--L $F_1$. }
\label{tab:amz_review}
% \vskip 0.15cm
\centering
\begin{small}

\begin{tabular}{ccccccccc}
\toprule
\multirow{2}{*}{Method} 
  & \multirow{2}{*}{Uncompressed} 
  & \multirow{2}{*}{SelectiveContext} 
  & \multirow{2}{*}{LLMLingua2}
  & \multirow{2}{*}{LTSC}
  & \multicolumn{3}{c}{\textbf{$K$-Token Merging (Ours)}} \\
\cmidrule(lr){6-8}
  &   &   &   &   
  & 2-Token & 3-Token & 4-Token \\
\midrule
\emph{Accuracy} & \textbf{93.51}$^{\pm 0.03}$ & 91.30$^{\pm 0.01}$ & 91.44$^{\pm 0.12}$ & \underline{93.49}$^{\pm 0.05}$ & 92.66$^{\pm 0.18}$ & 92.48$^{\pm 0.22}$ & 91.39$^{\pm 0.39}$ \\
    \emph{Length Reduction} & 0.0 & 44.6 & 51.0 & 0.1 & 50.0 & \underline{66.7}  & \textbf{75.0} \\
    \emph{P--L $F_1$} & 0.000 & 0.599 & 0.655 & 0.002 & 0.650 & \underline{0.775} & \textbf{0.824}  \\
\bottomrule
\end{tabular}

\end{small}
\vskip -0.1 cm
\end{table*}

In practice, different methods may trade off performance and length reduction differently; a method with stronger compression may exhibit slightly lower performance, and vice versa. To quantitatively evaluate this trade-off, we propose the \textbf{Performance--Length Reduction $F_1$ score (P--L $F_1$)}, defined as the harmonic mean of performance and length reduction, analogous to the classical $F_1$ score.

The P--L $F_1$ score is computed as:
\begin{equation}
    \text{P--L } F_1 = \frac{2 P L}{P + L},
\end{equation}
where $P \in [0,1]$ denotes the normalized performance score and $L$ denotes the length reduction ratio.

For the Textualized Tree and Amazon Reviews datasets, we directly use classification accuracy as $P$. For CommitPackFT, we use a \emph{relative perplexity ratio}. Let $\text{ppl}_i$ be the perplexity of method $i$ and $\text{ppl}_{\min} = \min\limits_i \text{ppl}_i$. The normalized performance $P_i \in [0,1]$ for each method $i$ is given by:
\begin{equation}
    P_i = \frac{\text{ppl}_{\min}}{\text{ppl}_i},
\end{equation}
which ensures that lower perplexity corresponds to higher normalized performance. 

{Note that P--L $F_1$ is a summary metric that balances task performance and compression ratio, and may therefore favor a particular trade-off between the two. We thus use it only as supplementary evidence and base our main conclusions primarily on the raw task metrics and the Pareto frontier.}

% \begin{tcolorbox}[colback=white,colframe=black]

% \end{tcolorbox}

\subsection{Implementation Details}
% We use Qwen-2.5 0.5B \citep{qwen2.5} as the base model for all baselines. We apply LoRA- finetune to this base model separately on compressed prompts provided by different baselines, using the same LoRA module structure throughout (rank $r=4$, LoRA alpha $\alpha=16$, and drop out rate $0.05$). For the Tree Classification task, we adopt a curriculum learning strategy, where the model is initially trained on smaller trees and move to larger trees only when the performance cross a threshold. This stage involves approximately 19.6 to 34.6 million auxiliary data samples. For all other tasks, we train the model directly without such strategy. For the encoder of our model, we use a three layer MLP. All models are trained with AdamW optimizer and learning rate $1e-4$. 

{We mainly use Qwen-2.5 0.5B \citep{qwen2.5} as the base model for all baselines. We also evaluate our method on the larger Qwen-2.5 Coder 7B \citep{qwen2.5-coder} using CommitPackFT.} We apply LoRA finetuning to this base model separately on the compressed prompts produced by different baselines, using the same LoRA configuration across all experiments (rank $r=4$, LoRA scaling factor $\alpha=16$, and dropout rate $0.05$. We apply LoRA adaptation across all layers, including both the self-attention modules and the MLP components).

For the Tree Classification task, we adopt a curriculum learning strategy: the model is first trained on smaller trees and progresses to larger trees only after its performance surpasses a predefined threshold. This stage uses approximately 19.6--34.6 million auxiliary training samples. For all other tasks, the model is trained directly without curriculum learning.

The encoder in our model is implemented as a three-layer MLP. All models are trained using the AdamW~\citep{loshchilov2017decoupled} optimizer with a learning rate of $1\times10^{-4}$.

For results with repeated runs, we report the mean and sample standard deviation. Length reduction ratios and the P--L $F_1$ scores derived from the mean performance are reported once.

\begin{algorithm}[t]
\begin{mdframed}
\SetAlgoLined
\SetKwProg{Class}{class}{:}{}
\SetKwProg{Method}{def}{:}{}
% \SetKwProg{Init}{def init}{:}{}
\SetKw{Self}{self}

\tcp{Encoder compressing k embeddings into 1 embedding}
\Class{Average-Initialized Encoder}{
    \BlankLine
    \Method{init(\Self, embedding\_dim: integer, k: integer)}{
        \Self.net $\leftarrow$ MLP(input\_dim = k * embedding\_dim, output\_dim = embedding\_dim)\;
    }
    
    \BlankLine
    \Method{forward(\Self, x: matrix)}{
        \tcp{Input shape:(batch\_size, embedding\_dim, k)}
        mean $\leftarrow$ Mean(x, axis = -1)\;

        x\_flat $\leftarrow$ Reshape(x, shape = (batch\_size, embedding\_dim * k))\;
        
        output $\leftarrow$ mean + \Self.net(x\_flat)\;
        
        \Return{output}\;
    }
}
\BlankLine

\caption{Pseudo Code for Average-Initialized Encoder. Since the MLP is initialized near 0, this encoder at first outputs a compressed embedding that closely matches the average-pooled embedding of the $K$ tokens, giving training a stable starting point.}
\label{alg:avg_encoder}
\end{mdframed}
\end{algorithm}

\subsection{Result Discussion}

{\textbf{\emph{Textualized Tree}}. \tabref{tab:tree_classify} reports the classification accuracies and length reductions, which are also visualized in \figref{fig:expr_result}(a). Relative to the 99.98\% uncompressed accuracy, 2-, 3-, and 4-Token Merging achieve mean accuracies of 99.79\%, 98.73\%, and 98.46\% while reducing input length by 50.0\%, 66.7\%, and 75.0\%, respectively. Thus, even the most aggressive setting loses only 1.52 percentage points of accuracy.}

{All three $K$-Token Merging configurations lie on the Pareto frontier: increasing $K$ provides progressively greater compression with a modest accuracy tradeoff. In comparison, LLMLingua2 reaches only 82.17\% accuracy at 27.0\% length reduction, while LTSC preserves 99.68\% accuracy but reduces length by only 27.1\%. Aligned with this conclusion, the 4-Token model obtains the highest P--L $F_1$ score of 0.851.}

\begin{table*}
\setlength{\tabcolsep}{2pt}
\caption{Perplexity, Length Reduction Ratio (\%), and P--L $F_1$ on CommitPackFT. We \textbf{bold} the best result and \underline{underline} the second-best result for each metric. On P--L $F_1$, our method obtains both the best and the second-best performance.}
\label{tab:commitpackft}
\vskip -0.1cm
\centering
\begin{small}

\begin{tabular}{ccccccccc}
\toprule
\multirow{2}{*}{Method} 
  & \multirow{2}{*}{Uncompressed} 
  & \multirow{2}{*}{SelectiveContext} 
  & \multirow{2}{*}{LLMLingua2} 
  & \multirow{2}{*}{LTSC}
  & \multicolumn{3}{c}{\textbf{$K$-Token Merging (Ours)}} \\
\cmidrule(lr){6-8}
  &   &   &   &   
  & 2-Token & 3-Token & 4-Token \\
\midrule
\emph{Perplexity} & \textbf{1.293}$^{\pm 0.001}$ & 1.380$^{\pm 0.000}$ & 1.381$^{\pm 0.000}$ & \underline{1.296}$^{\pm 0.000}$ & 1.315$^{\pm 0.024}$ & 1.346$^{\pm 0.031}$ & 1.379$^{\pm 0.012}$ \\
    \emph{Length Reduction} & 0.0  & 39.9 & 30.0 & 17.2 & 50.0 & \underline{66.7}  & \textbf{75.0}\\
    \emph{P--L $F_1$}	 & 0.000  &  0.560 &	0.454	& 0.293 & 0.663 &	\underline{0.787} &	\textbf{0.833}\\
\bottomrule
\end{tabular}

\end{small}
\vskip -0.3cm
\end{table*}

% \tabref{tab:tree_classify} presents results for the Textualized Tree, which is also visualized in \figref{fig:expr_result} (a). 
% Our $K$-Token merging models achieve high accuracies of 99.91\%, 98.63\%, and 98.38\%, with the 4-Token merging model attaining the highest Length Reduction Ratio of 75\% and the highest P--L $F_1$ score of $0.851$, surpassing the strongest baseline by 28.2\%. Hard prompt compression methods suffer from information loss. For example, LLMLingua2 reduce the input length by only 27\% and suffer a performance drop to 82.17\%, which gives it a low P-C $F_1$ score of $0.406$. For soft prompt compression method LTSC, while effectively preserve the content information, it does not leverage the inefficiency in the latent embedding space, which results in a relative low Length Reduction Ratio of 27.1\%. The uncompressed baseline achieves the highest accuracy of 99.97\%. The close accuracy of our $K$-Token models to the uncompressed baseline suggests that our method successfully preserve the input information while maintaining a high Length Reduction Ratio.

{\textbf{\emph{Amazon Reviews}}. \tabref{tab:amz_review} and \figref{fig:expr_result}(b) show the accuracy-compression tradeoff on natural-language sentiment classification. The uncompressed model reaches 93.51\% accuracy. With \mbox{2-,} 3-, and 4-Token Merging, accuracy remains at 92.66\%, 92.48\%, and 91.39\% while input length is reduced by 50.0\%, 66.7\%, and 75.0\%, respectively. In particular, the 2-Token model loses only 0.85 percentage points while halving the input length.}

{Our three configurations lie on the Pareto frontier, showing that each additional level of compression yields a distinct performance--length tradeoff. LTSC nearly matches the uncompressed accuracy at 93.49\% but reduces length by only 0.1\%. SelectiveContext and LLMLingua2 obtain more than 40\% reduction, but their accuracies remain below those of our 2- and 3-Token models. The 4-Token model obtains the highest P--L $F_1$ score of 0.824, indicating a favorable balance between accuracy and compression.}

% For this natural language dataset, soft prompt compression method LTSC struggles to find structured, repeatable patterns in corpus, resulting in a Length Reduction Ration of only 0.1\%. Hard prompt compression method are more suited for this scenario due to the sparsity of key information in the corpus, with both achieve over 40\% Length Reduction Ratio and maintain an Accuracy over 90\%.

% However, all these methods are inferior to our method in terms of P--L $F_1$ score. our $K$-Token Merging models maintain high accuracy, with the 2-Token model performing best among them at 92.51\%, just 1.03\% lower than the uncompressed baseline, while achieving a 50\% input length reduction. The 4-Token model, with the highest compression ratio of 75.0\%, achieves the highest P--L $F_1$ score of $0.822$, 25.5\% over the strongest baseline.

{\textbf{\emph{CommitPackFT}}. \tabref{tab:commitpackft} and \figref{fig:expr_result}(c) present the perplexity--compression tradeoff on code editing, where lower perplexity is better. The uncompressed model obtains a perplexity of 1.293. The 2-, 3-, and 4-Token models obtain perplexities of 1.315, 1.346, and 1.379 while reducing input length by 50.0\%, 66.7\%, and 75.0\%, respectively. These correspond to relative perplexity increases of only 1.7\%, 4.1\%, and 6.7\%.}

{All three of our configurations lie on the Pareto frontier. By contrast, SelectiveContext and LLMLingua2 have higher perplexity than our 2-Token model while providing less length reduction, and LTSC obtains near-uncompressed perplexity (1.296) but reduces length by only 17.2\%. The perplexities and Pareto dominance therefore prove that $K$-Token Merging offers a favorable tradeoff on code. As a result, the 4-Token model achieves the highest P--L $F_1$ score of 0.833.}

\subsection{Scaling to a Larger Model}
\label{sec:larger_model}

{The main experiments use a 0.5B-parameter backbone. To examine whether the runtime benefits of token merging persist at a larger scale, we additionally evaluate 2-Token Merging with a Qwen Coder 7B model on CommitPackFT. Table~\ref{tab:larger_model} compares it with the corresponding uncompressed model.}

\begin{table}[!htbp]
\centering
\setlength{\tabcolsep}{4pt}
\caption{Performance and inference efficiency on the Qwen Coder 7B model. Lower perplexity, latency, and peak memory are better; higher throughput is better. The best value for each metric is bolded.}
\label{tab:larger_model}
\resizebox{\columnwidth}{!}{%
\begin{tabular}{lcc}
\toprule
Metric & No Compression & 2-Token Merge \\
\midrule
Perplexity $\downarrow$ & \textbf{1.263} & 1.277 \\
Latency (s) $\downarrow$ & 0.789$^{\pm 0.151}$ & \textbf{0.624}$^{\pm 0.139}$ \\
Throughput (samples/s) $\uparrow$ & 5.271$^{\pm 1.138}$ & \textbf{6.736}$^{\pm 1.577}$ \\
GPU peak memory (MB) $\downarrow$ & \textbf{35{,}983}$^{\pm 1{,}026}$ & 38{,}349$^{\pm 938}$ \\
\bottomrule
\end{tabular}%
}
\end{table}

{Relative to the uncompressed model, 2-Token Merging increases perplexity only slightly, from 1.263 to 1.277 (a 1.1\% relative increase), while reducing latency by 20.9\% and improving throughput by 27.8\%. Peak memory usage increases by 6.6\% in this setting. Due to limited computational resources, we evaluate with a batch size of only 2, under which GPU memory usage is dominated by the base model and our additional encoder, leading to the modest increase in peak memory. With larger batch sizes, however, the memory savings from token compression are expected to become more pronounced, as reducing the number of processed tokens can offset the fixed memory overhead introduced by the encoder.}
\subsection{Ablation Studies}
\label{expr:ablation}

\paragraph{Component ablation.}
{To isolate the contribution of each component, we conduct a controlled ablation on CommitPackFT. We compare full $K$-Token Merging (KTM), mean pooling without the learnable encoder residual, and an uncompressed control with a matched extra-parameter budget. The last configuration uses a comparable encoder that maps $K$ input embeddings to $K$ output embeddings, thereby retaining the original sequence length. For this focused comparison, all configurations are trained for 10 epochs.}

\begin{table}[!htbp]
\centering
\setlength{\tabcolsep}{3pt}
\caption{Component ablation on CommitPackFT after 10 epochs. The best value for each metric is bolded, and the second-best value is underlined.}
\label{tab:component_ablation}
\resizebox{\columnwidth}{!}{%
\begin{tabular}{lccc}
\toprule
Metric & KTM (Ours) & Mean Pooling Only & No Compression \\
\midrule
Perplexity $\downarrow$ & \underline{1.447} & 1.453 & \textbf{1.364} \\
Latency (s) $\downarrow$ & \underline{0.358}$^{\pm 0.103}$ & \textbf{0.353}$^{\pm 0.102}$ & 0.546$^{\pm 0.118}$ \\
Throughput (samples/s) $\uparrow$ & \underline{17.792}$^{\pm 4.124}$ & \textbf{18.038}$^{\pm 4.213}$ & 11.377$^{\pm 2.148}$ \\
GPU peak memory (MB) $\downarrow$ & \underline{9{,}552}$^{\pm 1{,}314}$ & \textbf{9{,}232}$^{\pm 1{,}314}$ & 13{,}006$^{\pm 1{,}444}$ \\
\bottomrule
\end{tabular}%
}
\end{table}

{As shown in Table~\ref{tab:component_ablation}, adding the learnable residual improves perplexity from 1.453 to 1.447 relative to mean pooling alone, with only small differences in latency, throughput, and memory. Compared with the matched uncompressed control, full KTM reduces latency by 34.4\%, increases throughput by 56.4\%, and lowers peak memory by 26.6\%, at the cost of a slightly higher perplexity (+6.09\%). These results indicate complementary roles for the two components: the learnable encoder residual recovers semantic quality beyond fixed mean pooling, while reducing the sequence length provides the primary efficiency gains.}

\paragraph{Embedding initialization strategies ablation.}

\begin{figure}[!htbp] %{wrapfigure}{r}{0.45\columnwidth}
\vskip -0.29in
\centering
\includegraphics[width=0.900\columnwidth]{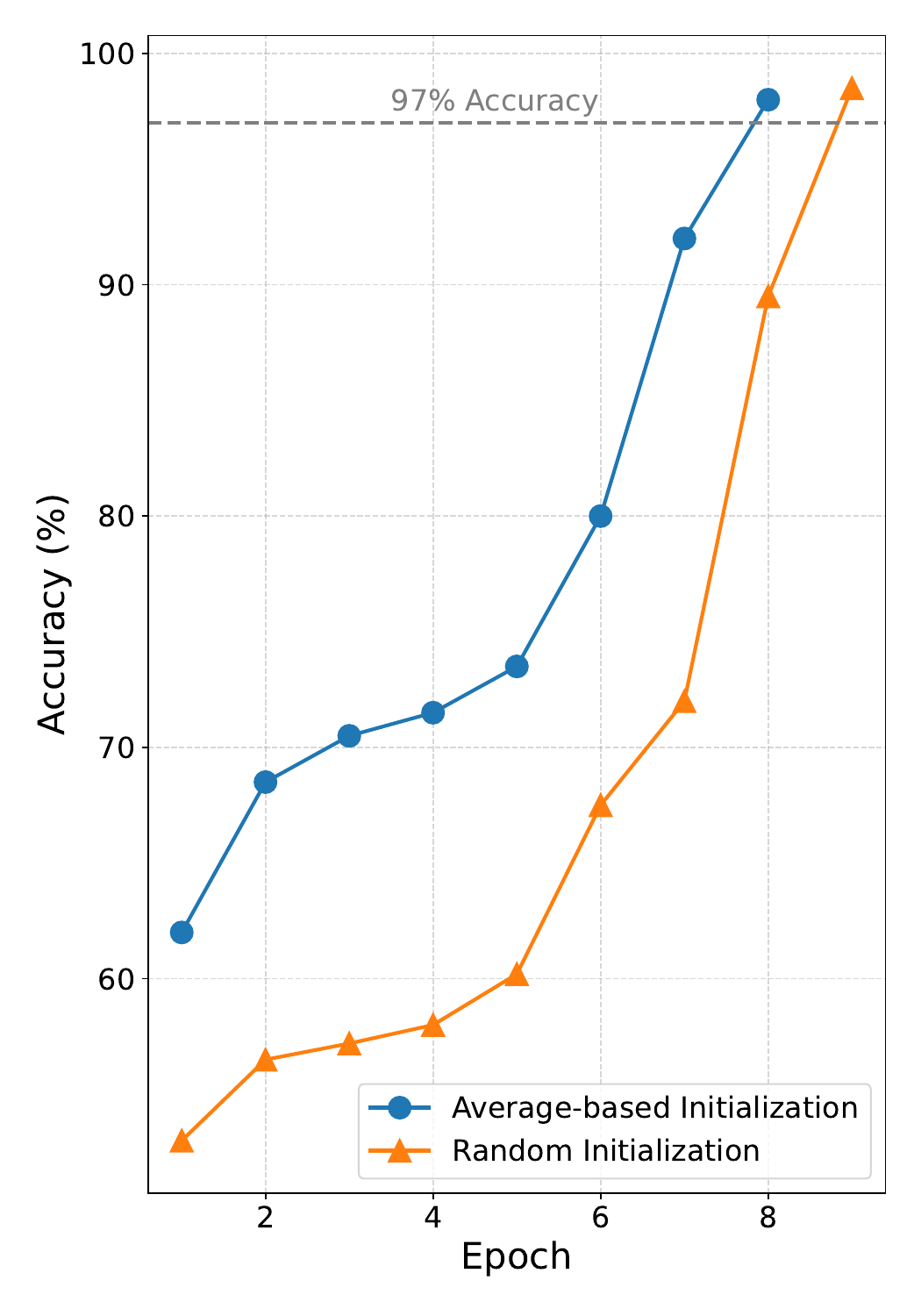}
\vskip -0.10in
\caption{
Ablation Study on Embedding Initialization Strategies. We evaluate both initialization strategies on the Textualized Tree classification task using smaller trees (only 5 nodes). \textbf{All experiments use our 4-Token Merging Model}. The results show that the model with average-based initialization converges one epoch faster than the model with random initialization when reaching 97\% accuracy. %  (obtained by average pooling the original token embeddings) 
}
\label{fig:ablation}
\vskip -0.1in
\end{figure}

As shown in Fig. \ref{fig:ablation}, we present an ablation study comparing different embedding initialization strategies for the compressed embeddings. Our average-based initialization achieves 97\% classification accuracy in just 8 epochs -- one epoch faster than random initialization. This indicates that the average-based approach enables faster convergence and serves as a more effective initialization method than random initialization.

\subsection{Case Study}
We include a case study of our $2$-Token Merging model trained on the CommitPackFT dataset to demonstrate that, even with compressed inputs, the model can still preserve key information and follow instructions to generate correct outputs.

\begin{lstlisting}[language=Python, caption={Code input used for our model's case study.}, label={lst:before}]
import math
import matplotlib.pyplot as plt
import numpy as np

# Define the function
def compute_sin(x):
    return np.sin(x)

# Generate input values
x_vals = np.linspace(-2 * np.pi, 2 * np.pi, 500)
y_vals = compute_sin(x_vals)

# Plotting
plt.figure(figsize=(8, 4))
plt.plot(x_vals, y_vals, label='sin(x)')
plt.show()
\end{lstlisting}

In this example, we provide the original code in Listing \ref{lst:before} and instruct the model to ``plot the `cos' function at the same time.''

The generated code is shown in Listing \ref{lst:after}. Two observations are noteworthy. First, the model successfully generates an additional function to plot the cosine curve. Second, rather than introducing a new variable, it reuses the previously defined variable ``x\_vals'' (highlighted in yellow) as the function domain, ensuring that both plots share the same range. 

These two points indicate that, despite compressing both the original code and the instruction, the model can still recover most relevant information and generate the correct implementation. This suggests our method effectively leverages redundancy in the embedding space while preserving model performance.
% These two points indicate that, despite the compression applied to both the original code and the instruction, the model is still able to recover most of the relevant information and produce the correct implementation. This result suggests that our method effectively leverages redundancy in the embedding space while maintaining model performance.

% \subsection{Implementation Details of Average Initialization}

% We include a pseudo code of our average-based encoder initialization, as is shown in Alg. \ref{alg:avg_encoder}. The core idea is to let the output to be a residual combination of the mean of k original embeddings and the output of a multi-layer perceptron. In practice, we random initialized the weights of the multi-layer perceptron for better convergence speed.

% We provide the pseudocode for our average-based encoder initialization in Algorithm \ref{alg:avg_encoder}. The core idea is to produce an output as a residual combination of the mean of $k$ original embeddings and the output of a multi-layer perceptron. In practice, we randomly initialize the weights of the MLP to improve convergence speed.

\section{Future Directions}

 \textbf{Compression During Generation:} Currently, we do not apply token compression to generated tokens for simplicity. Exploring how to extend our compression technique to the generation phase could lead to significant memory and efficiency gains.  

 \textbf{Recursive Encoding:} In our setup, we apply the encoder once to compress the input tokens. A natural extension is to apply the encoder recursively -- e.g., compress 8 tokens into 4, then compress those 4 into 2 -- enabling progressively shorter representations. This hierarchical compression approach may further improve efficiency while preserving performance.
 
\begin{lstlisting}[float=tbp, language=Python, label={lst:after}, caption={Output code. Our model follows the instructions to add an additional plot of cosine function. It  reuse the previously defined variable ``x\_vals'' (highlighted in yellow) as the function domain. }, escapeinside={(*@}{@*)}]
import math
import matplotlib.pyplot as plt
import numpy as np

# Define the function
def compute_sin(x):
    return np.sin(x)

# Generate input values
(*@\colorbox{yellow}{x\_vals}@*) = np.linspace(-2 * np.pi, 2 * np.pi, 500)
y_vals = compute_sin(x_vals)

# Plotting
plt.figure(figsize=(8, 4))
plt.plot(x_vals, y_vals, label='sin(x)')
plt.show()

# Plotting the "cos" function
plt.figure(figsize=(8, 4))
plt.plot((*@\colorbox{yellow}{x\_vals}@*), np.cos(x_vals), label='cos(x)')
plt.legend()
plt.show()
\end{lstlisting}

\textbf{Adaptive Compression:} In this paper, we apply uniform compression over every fixed group of $K$ tokens, regardless of the statistical properties of different $K$-grams. An adaptive strategy -- such as compressing 2 tokens in high-frequency regions and 16 in low-frequency ones -- could further improve the compression ratio while maintaining or even boosting performance.

% \section{Future Work}

% There are several promising directions for future exploration:

% \begin{enumerate}
%     \item \textbf{Compression During Generation:} Currently, our method employs standard autoregressive generation, which prevents the application of token compression to generated tokens. Exploring how to extend our compression technique to the generation phase could lead to significant memory and efficiency gains.
    
%     \item \textbf{Recursive Encoding:} In our current setup, we apply the encoder once to compress the input tokens. A natural extension is to apply the encoder recursively--e.g., compress 8 tokens into 4, then compress those 4 into 2--enabling progressively shorter representations. This hierarchical compression approach may further improve efficiency while preserving performance.
    
%     \item \textbf{Adaptive Compression:} At present, we apply uniform compression over every fixed group of $k$ tokens, regardless of the statistical properties of different $k$-grams. An adaptive strategy---such as compressing 2 tokens in high-frequency regions and 16 in low-frequency ones---could improve the compression ratio while maintaining or even boosting performance.
% \end{enumerate}

\section{Conclusions}

In this paper, we introduced $K$-Token Merging, a latent-space compression framework that merges groups of tokens into single embeddings to reduce the effective input length of LLMs. By combining a lightweight encoder with LoRA adaptation, our method enables LLMs to process compressed inputs while preserving standard generation. Experiments across synthetic, natural-language, and code tasks demonstrate up to 75\% input length reduction with limited performance degradation. Together, these results suggest that exploiting redundancy in the embedding space is a promising direction for improving the efficiency of long-context LLM inference.

\section*{Limitations}

Our work has the following limitations that we leave for future research:

\textbf{No compression during generation.} Our method compresses only input tokens and does not compress generated tokens, limiting efficiency gains as the output sequence grows.

\textbf{Fixed compression ratio.} We apply uniform compression to every group of $K$ tokens regardless of corpus statistics. Adaptive strategies (e.g., smaller $K$ in dense regions and larger $K$ in sparse ones) could further improve compression while maintaining performance.

\textbf{Limited model coverage.} Most experiments use Qwen 2.5 0.5B, with an additional scaling experiment on Qwen Coder 7B. Evaluation across more model families, parameter scales, and architectures is needed to establish broader generality.

% \section*{Acknowledgments}

% This document has been adapted
% by Steven Bethard, Ryan Cotterell and Rui Yan
% from the instructions for earlier ACL and NAACL proceedings, including those for
% ACL 2019 by Douwe Kiela and Ivan Vuli\'{c},
% NAACL 2019 by Stephanie Lukin and Alla Roskovskaya,
% ACL 2018 by Shay Cohen, Kevin Gimpel, and Wei Lu,
% NAACL 2018 by Margaret Mitchell and Stephanie Lukin,
% Bib\TeX{} suggestions for (NA)ACL 2017/2018 from Jason Eisner,
% ACL 2017 by Dan Gildea and Min-Yen Kan,
% NAACL 2017 by Margaret Mitchell,
% ACL 2012 by Maggie Li and Michael White,
% ACL 2010 by Jing-Shin Chang and Philipp Koehn,
% ACL 2008 by Johanna D. Moore, Simone Teufel, James Allan, and Sadaoki Furui,
% ACL 2005 by Hwee Tou Ng and Kemal Oflazer,
% ACL 2002 by Eugene Charniak and Dekang Lin,
% and earlier ACL and EACL formats written by several people, including
% John Chen, Henry S. Thompson and Donald Walker.
% Additional elements were taken from the formatting instructions of the \emph{International Joint Conference on Artificial Intelligence} and the \emph{Conference on Computer Vision and Pattern Recognition}.

% Bibliography entries for the entire Anthology, followed by custom entries
%\bibliography{anthology,custom}
% Custom bibliography entries only
\bibliography{custom}

\appendix

% This is an appendix.

\end{document}